%% file: main.tex
\documentclass{article} 
\usepackage[final]{neurips_2025}
\usepackage{times}

\input{math_commands.tex}

\usepackage{url}
\usepackage{todonotes}
\usepackage[acronym]{glossaries}
\usepackage{chemformula}
\usepackage{graphicx}
\usepackage{hyperref}
\usepackage{cleveref}
\usepackage{subcaption}
\usepackage{verbatim}
\usepackage{colortbl}
\usepackage{array}
\usepackage{tcolorbox}

\newcommand{\astar}[0]{A$^*$ }
\newcommand{\astarNS}[0]{A$^*$}

\newacronym{ML}{ML}{machine learning}
\newacronym{LLM}{LLM}{Large Language Model}
\newacronym{DFT}{DFT}{Density Functional Theory}
\newacronym{dspy}{DSPy}{Declarative Self-improving Python}
\newacronym{SME}{SME}{subject matter expert}
\newacronym{MCTS}{MCTS}{Monte Carlo Tree Search}
\newacronym{SI}{SI}{Supplementary Information}

\title{Coupling Language Models with Physics-based Simulation for Synthesis of Inorganic Materials}

\author{Edward W. Staley, Tom Arbaugh, Michael Pekala, Alexander New, Christopher D. Stiles, Nam Q. Le \\
Johns Hopkins Applied Physics Laboratory, Laurel, MD 20723\\
\texttt{\{\textit{firstname.lastname}\}@jhuapl.edu} \\
\And
Gregory Bassen, Wyatt Bunstine, Tyrel McQueen \\
Johns Hopkins University, Baltimore, MD 21218 \\
\texttt{mcqueen@jhu.edu}
}

\makeatletter
\def\@trackname{}
\makeatother

\begin{document}

\maketitle

\begin{abstract}

Modern generative \gls{ML} models can propose novel inorganic crystalline materials with targeted properties; however, synthesis planning of these materials remains difficult due to the complexity of the associated physical processes and limited availability of computational tools. We introduce a novel hybrid framework to evaluate \glspl{LLM} in inorganic synthesis planning by combining thermodynamic databases with simplified kinetics models to approximate realistic synthesis conditions. As a case study, we focus on the niobium–oxygen system, which features multiple industrially relevant oxide phases with well-characterized data. In computational simulations, we compare \gls{LLM}-generated synthesis routes with classical path-planning algorithms, showing that the implicit priors in \glspl{LLM} can yield more viable strategies. In our evaluation setting, classical search methods serve primarily as a foil rather than a direct competitor. This illustrates the relative complexity of the problem and highlights where the LLM’s implicit priors add value.

\end{abstract}

\section{Introduction}

The discovery and synthesis of novel inorganic materials underpin advances in energy technologies, electronics, catalysis, and structural applications. Generative \gls{ML} models have emerged as powerful tools for proposing candidate materials with targeted properties (\cite{wilfong2025ternary}). Such models can sample vast material design spaces, but the subsequent step of translating a computationally designed material into a feasible synthesis plan presents a significant bottleneck. Traditional approaches for designing synthesis plans, such as exploring energy landscapes, lack reliable methods for determining reaction coordinates and are typically based on a combination of literature sources, intuition, and scientist experience.  

We present a preliminary evaluation of LLMs for inorganic synthesis planning using a hybrid framework that augments thermodynamic databases with simplified kinetic models. As a case study, we investigate the binary niobium–oxygen system, a chemically rich space with multiple industrially relevant oxide phases and well-characterized thermodynamics.
By comparing LLM-generated synthesis routes with those from a classical path-planning algorithm, we assess the extent to which the embedded scientific priors in LLMs can improve planning quality. 

\section{Related Work}


\glspl{LLM}, trained on vast corpora spanning scientific literature and general text, offer rich possibilities to augment synthesis planning tools. 

Outside of materials applications, there are works that consider the suitability of current generative models for reasoning through problems more generally~\citep{shojaee2025illusion}. 
The ability of LLMs to encode prior knowledge raises the possibility of them generating grounded synthesis pathways, and a number of works have considered leveraging generative models to evaluate or determine synthesis pathways. \cite{pu2024leveraging} assessed LLMs in terms of responding to text-based knowledge prompts focusing on synthesis of gold nanoparticles. More direct prediction of synthesis routes, in terms of identification of precursors and conditions, was performed by \cite{prein2025language}, including benchmarking against datasets of specific reactions in solid-state synthesis. In this work, we propose a complementary approach using approximate but physically-grounded thermodynamic simulations, rather than sets of reaction-specific data, for quantitative assessment of proposed routes across a broad material system.

\section{Methods} \label{methods}

\subsection{Material System and Problem Statement} \label{sec:problemstatement}
Assessment using physics-based simulation in this work relies on the availability of well-validated thermochemical data. We use the binary \ch{Nb-O} system as a case study; phase diagrams are shown in~\cref{fig:phasediagrams} in the \gls{SI}. This system exhibits several competing solid oxide phases which allow us to pose challenges of varied complexity.
Benchmarking synthesis models in well-understood systems, such as \ch{Nb-O}, will prepare for future extensions to systems without completely assessed phase diagrams, which is ultimately desired for novel materials.

\label{sec:methods-simulator}


To simulate the kinetics of the system, we adopt a Johnson–Mehl–Avrami (JMA) style approach~\cite{kinetics} in which phase fractions evolve according to $f(\Delta t) = (f_{\mathrm{eq}} - f_{0}) \left[ 1 - \exp(-k \Delta t) \right]$, where $f_0$ and $f_{\mathrm{eq}}$ are the initial and equilibrium phase fractions, and $k$ is the rate constant given by $k = k_0 \exp\left[ -E_{a}/(RT) \right]$, with activation energy $E_a$, gas constant $R$, and temperature $T$.


Equilibrium phase fractions were computed using the CALPHAD technique with PyCalphad~\cite{Otis2017pycalphad} and the thermodynamic database of~\cite{tdb}, at 10 K intervals from 300 K to 3000 K, with a compositional spacing of 1 at.\%. We set $E_{a} = 95.4$~kJ/mol, based on the experimentally-reported activation energy for diffusion-limited oxidation of Nb~\cite{diffusion,oxidation}, and $k_{0} = 100$~s$^{-1}$ to reproduce the reported timescale of the transformation of \ch{Nb} to \ch{Nb2O5} at 900 K~\cite{diffusion}. 

\Cref{tbl:challenge} shows the four challenge problems that we assessed.
Each problem consists of a starting state and a desired goal state, where ``state'' is defined as a convex combination of realized phases in the material system together with an associated temperature (pressure is assumed fixed at 1 bar).
These problems were designed to have a range of difficulty. They vary in features such as requiring temperature variation, single or multiple phases, and/or traversal of ``intermediate'' phases in the phase diagram. \Cref{tbl:challenge-detailed} in the \gls{SI} provides further details on the challenges.

\begin{table}
\centering
\caption{The four challenge problems assessed using our physics-based simulator.}
\begin{tabular}{lllll}
ID & \multicolumn{2}{l}{Start State} & \multicolumn{2}{l}{Goal State} \\
\hline
    1 & 300~K & 100\% \ch{Nb} & 300~K & 100\% NbO \\
    2 & 300~K & 100\% \ch{Nb} & 2200~K & 50\% NbO, 50\% liquid \\ 
    3 & 300~K & 100\% \ch{Nb} &  300~K & 50\% \ch{NbO2} 50\% \ch{Nb2O5} \\
    4 & 300~K & 100\% \ch{Nb2O5} & 300~K & 100\% \ch{NbO} 
\end{tabular}
\label{tbl:challenge}
\end{table}

In response to each challenge, a respondent must provide a synthesis ``recipe,'' which may consist of any sequence of three options: add precursor, change temperature, or wait. ``Add precursor'' may choose from the two pure compounds: Nb or O$_2$. ``Change temperature'' may set any temperature between 300~K and 3000~K. ``Wait'' holds the temperature and composition constant while allowing phase fractions to evolve according to thermodynamics and kinetics modeled by the simulator.

Executing the synthesis recipe results in some final state which is then scored by comparing it to the problem goal state.  Any solution that does not achieve the final temperature to within 20~K is deemed unresponsive and assigned the maximum error of 1.0. Otherwise, the solution is scored by computing a scaled mean squared error between the goal and realized phase fractions. In the future, other scoring methodologies might be considered, including adding intermediate constraints (such as avoiding a prohibited intermediate phase) or taking more careful account of the geometry of the simplex when evaluating the phase differences.

\subsection{Candidate Synthesis Planning Models}

We compare two fundamentally different approaches for solving the synthesis planning problem: (1) as a classical path planning problem to help quantify problem difficulty and (2) by leveraging an \gls{LLM} to explore the inclusion of prior knowledge.
There are a variety of classical approaches (e.g. ~\cite{Kocsis2006mcts}) that treat the problem as a search space. We elected to utilize \astar search~\cite{astarpaper} for its simplicity and because precision to within a small percentage of the goal was acceptable. The search heuristics of \astar are illustrative of techniques that can be guided by mathematical expressions but are hard to shape with abstract priors or domain knowledge.
We discretize the state space into 100 bins along each of the phase fraction, element fraction, and temperature dimensions. 
For a found state in \astar, the cost to reach that state is taken to be the number of steps in the path. Our heuristic for estimating the remaining cost to the goal is the Euclidean distance between two normalized states.

In contrast to classical path planning, the \gls{LLM} has prior knowledge of materials synthesis and can be given further knowledge through prompting.
We also provide the \gls{LLM} with access to the simulator itself, which it can invoke as a ``tool'' in generating a solution.
This is analogous to domains like mathematics, where \glspl{LLM} are paired with tools (e.g., theorem provers) to derive new results. 

Note that while an LLM-guided synthesis search can incorporate prior knowledge, it is also more computationally expensive, and many-step LLM reasoning is an open area of research \cite{aghajohari2025markovianthinker, fei2023extendingcontextwindowlarge}. We instructed the LLM to continue attempting a problem until it reached a conclusion (\cref{fig:llmcontext}) and did not otherwise limit its generation length. We let \astar run until it found a solution or exhausted system resources while building the search graph (128GB RAM).





Our LLM is implemented as a ReAct-based agent~\cite{yao2023react} using the \gls{dspy} framework~\cite{khattab2024dspy,khattab2022demonstrate}, which we attach to OpenAI's GPT4o model~\cite{openai2024gpt4o}.
GPT4o was selected as a state-of-the-art non-reasoning model for which we could introduce the reasoning process ourselves through prompting.
Of course, future work would compare multiple models, including reasoning vs non-reasoning varieties.
Given that the performance of in-context learning depends upon the prompt ~\cite{min2022rethinking}, we also consider strategies for designing this prompt in more systematic ways.
For this preliminary study, we leveraged \gls{dspy}'s prompt tuning capabilities; in particular, its \texttt{LabeledFewShot} strategy which, given a training data set, randomly selects $k=4$ new examples to add to the prompt.\footnote{ 
Note \gls{dspy} includes more sophisticated optimization strategies that we propose to explore moving forward. We used random sampling as an initial proof-of-concept to determine whether prompt tuning seemed promising.}

\section{Results and Discussion}
\begin{figure}[b]
\centering
\vspace{-0.7cm}
\includegraphics[width=.75\textwidth,trim={0, 0, 10cm, 0},clip]{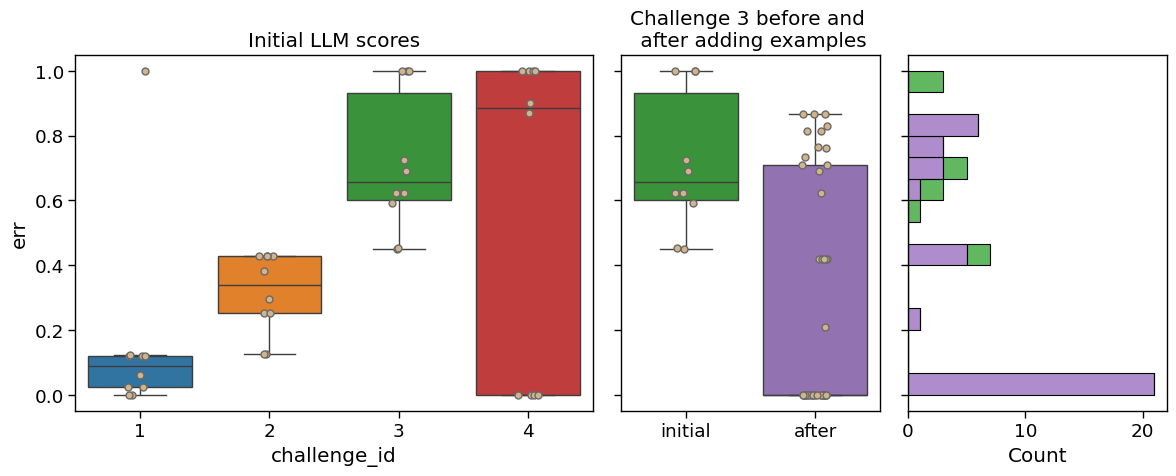}
\caption{\gls{LLM} performance on the 4 challenge problems (CP). 
(left) performance for the original, manually curated prompt; (right) performance on CP 3 before (green) and after (purple) prompt tuning.
We ran the prompt tuning procedure 5 times (with different seeds), hence there are more purple data points than green.
}
\label{fig:llmresults}
\end{figure}

Our primary results\footnote{
More details, including the best solutions found by the LLM for each challenge, are provided in the \gls{SI}.
}
are summarized in~\cref{tbl:resultscomparison} and~\cref{fig:llmresults}; the former shows a comparison of \astar and the \gls{LLM} agent on the challenge problems while the latter shows the full distribution of scores and provides a comparison of \gls{LLM} performance pre- and post-prompt tuning. 
In the case of~\cref{tbl:resultscomparison}, ``success'' is defined as the final state being within some approximate $\epsilon$-ball of the goal state; of course, other criteria might be considered in future work. 
Since the search space is deterministic, we only run \astar once, resulting in either 100\% success or failure. 
To account for the stochasticity of LLM generation, we run the LLM ten times on each problem.
In general, we find that the \gls{LLM} and \astar both exhibit some variability in performance across the four challenges.
The relative effectiveness of the \gls{LLM} suggests that its scientific priors indeed carry over to our simulation environment. 
Results from the preliminary prompt tuning (\cref{fig:llmresults}) suggests that more careful optimization is important to consider in future work.

\begin{table}[ht]
\centering
\caption{Comparison of \astar and \gls{LLM} (no prompt tuning) results. 
``States visited'' is the rough magnitude of state transitions simulated during the application of the method.
}
\label{tbl:resultscomparison}
\small
\begin{tabular}{l|cc|cc|cc|cc}
& \multicolumn{2}{c}{Challenge 1} & \multicolumn{2}{c}{Challenge 2} & \multicolumn{2}{c}{Challenge 3} & \multicolumn{2}{c}{Challenge 4}           \\
       &Success& States &Success& States &Success& States &Success& States            \\
Method & Rate & Visited & Rate & Visited & Rate & Visited & Rate & Visited            \\
\hline 
\rule{0pt}{2.5ex}
\astar     & 100\% & O($10^5$) & 0\% & $\ge$ O$(10^7$) & 0\% & $\ge$ O($10^7$) & 100\% & O($10^7$) \\
LLM    & 20\% & O(10) & 0\% & O(10) & 0\% & O(10)  & 40\% & O(10)                    
\end{tabular}
\end{table}

Challenges 1 and 4 were both solved by \astar\ and the LLM. However, \astar explored orders of magnitude more states than the LLM. Challenges 2 and 3 include a target with multiple phases, and was more difficult for both approaches. On challenge 2, the LLM obtained the correct phases, but not the correct fractions. On challenge 3, the LLM was only able to find solutions after prompt tuning. \astar  ran out of system resources before finding a solution to challenges 2 and 3.
A key challenge in applying \astar is that, while distance to the goal is primarily measured in phase fractions, the available simulation options only include elemental composition, temperature, and time. The relationship between these spaces is provided by the simulation itself and difficult to estimate a priori, making classic search strategies (even those leveraging heuristics) impractical. In contrast, the LLM is able to attempt sensible approaches for this domain by leveraging training data and prompt information, rather than an explicit notion of distance. Examples of LLM outputs are provided in \ref{fig:llmoutputsuccess} and \ref{fig:llmoutputfail}.

\section{Conclusions}

This work provides three main results.
First, we have demonstrated a strategy for using simple physics-based simulation as a virtual ``testbed'' to assess LLMs' ability to propose material synthesis plans. 
Second, we identified synthesis problems where the LLM's implicit priors appear to facilitate finding solutions more efficiently than naive search methods. 
Third, there are still very simple synthesis problems which both \astar and the example LLM fail to solve; however, this framework suggests straightforward ways to ground LLM predictions by invoking physics-based tools.
Both our simulation and our prompt optimization strategy are nascent and there are many opportunities to make this approach even more applicable to practical synthesis pathway planning.

\begin{ack}

EWS, TA, MP, CDS, and NQL gratefully acknowledge internal financial support from the Johns Hopkins Applied Physics Laboratory's Independent Research \& Development (IR\&D) Program for funding portions of this work. We would like to thank the reviewers for their valuable feedback.


\end{ack}

\newpage


\bibliographystyle{plainnat}
\bibliography{references}

\appendix
\section{Supplementary Information}

\subsection{Phase diagrams}

\begin{figure}[h!]
\centering
\begin{subfigure}[t]{0.49\textwidth}
\centering
\includegraphics[trim=0 120 0 50, width=\textwidth]{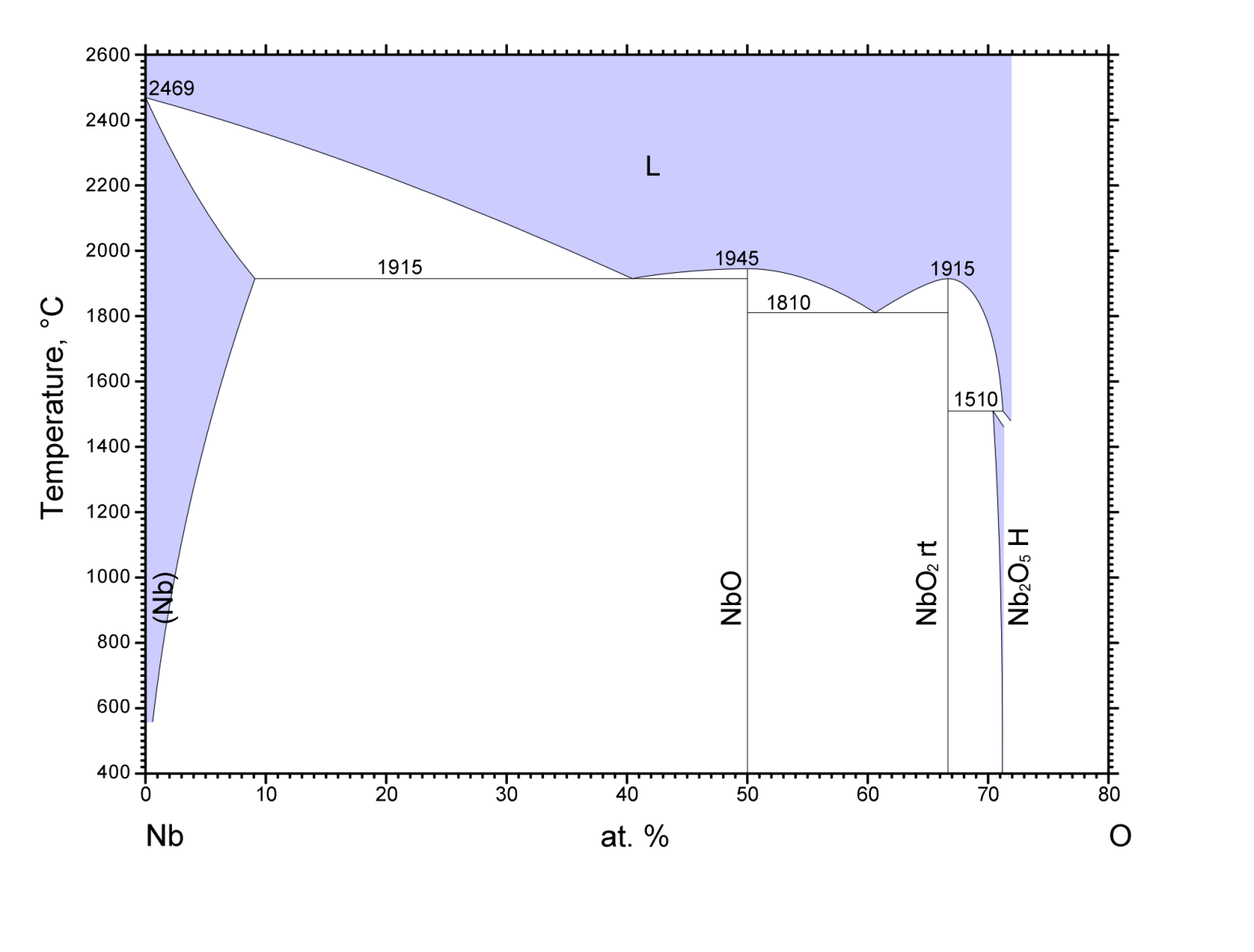}
\caption{Experimental.}
\label{fig:phase-experimental}
\end{subfigure}
\begin{subfigure}[t]{0.49\textwidth}
\centering
\includegraphics[width=\textwidth]{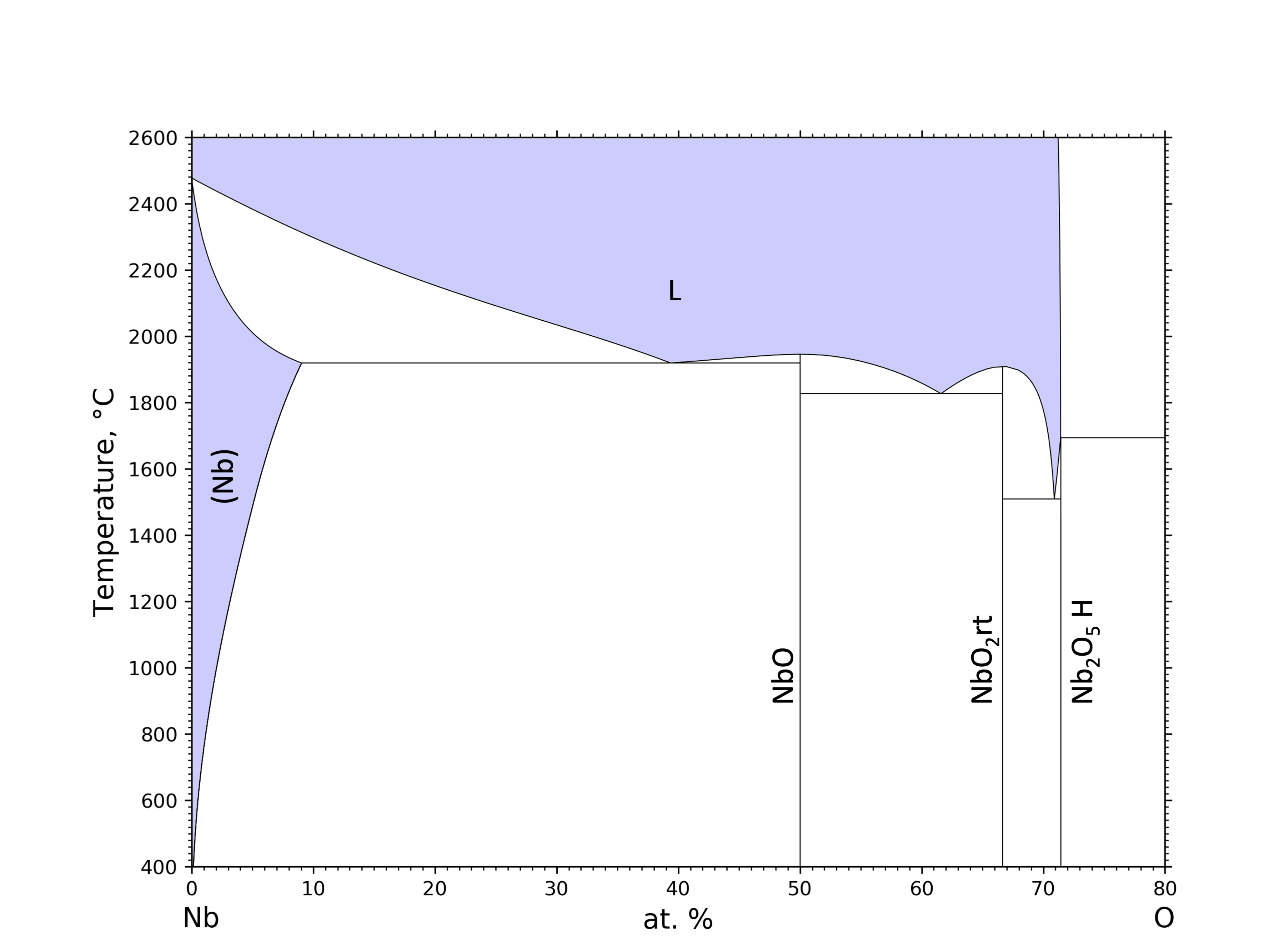}
\caption{Computational (CALPHAD).}
\label{fig:phase-calphad}
\end{subfigure}
\caption{
Phase diagrams for \ch{Nb-O}. Observe the good agreement between the experimental (a) and computational estimates (b) of the phases.
``L'' denotes the liquid phase, ``rt'' denotes the rutile \ch{NbO2} phase, ``H'' denotes the monoclinic \ch{Nb2O5} phase, and ``at.\%'' denotes atomic percent.
}
\label{fig:phasediagrams}
\end{figure}

For the binary \ch{Nb-O} system used in this work as a case study, two phase diagrams are shown in \cref{fig:phasediagrams}. The experimental phase diagram is reproduced from \cite{okamoto1990nb}, and we computed the phase diagram with the CALPHAD method as implemented in PyCalphad using the database of \cite{tdb}. The same database was used with PyCalphad to compute the phase fractions that would occur at thermodynamic equilibrium, which serve as input to the JMA-style approximation to the kinetics described in \Cref{sec:methods-simulator}.

\subsection{Details of challenge problems}

The four synthesis challenges used to assess \astar and LLMs in terms of their ability to generate synthesis plans are enumerated in \Cref{tbl:challenge-detailed}. Additionally, the search space for \astar normalized all temperatures by dividing by 3000K such that all dimensions of the search space fell within [0-1]. The phase fractions are naturally in this range without modification.

\begin{table}[th]
\small
\centering
\caption{Challenge problems related to the Nb-O oxide system posed to algorithms in this work.}
\label{tbl:challenge-detailed}
\begin{tabular}{r l l l p{1.7in}<{\raggedright}}
   & Start & Goal  & Example &  \\
ID & State & State & Solution & Comments \\
\hline
1  
& 
\begin{tabular}{@{}l@{}} 
$T_0 = 300$ K \\
100\% \ch{Nb} 
\end{tabular} 
&
\begin{tabular}{@{}l@{}} 
$T_f = 300$ K \\
100\% \ch{NbO} 
\end{tabular} 
&
\begin{tabular}{@{}l@{}} 
\texttt{add}(50 at.\% O) \\
\texttt{settemp}(1300 K) \\ 
\texttt{wait}(60 min) \\
\texttt{settemp}(300 K)
\end{tabular} 
& Direct composition change toward single oxide phase that is ``adjacent'' in composition space. \\
\hline
2
& 
\begin{tabular}{@{}l@{}} 
$T_0 = 300$ K \\
100\% \ch{Nb} 
\end{tabular} 
&
\begin{tabular}{@{}l@{}} 
$T_f = 2200$ K \\
50\% \ch{NbO}, 
50\% liquid 
\end{tabular} 
&
\begin{tabular}{@{}l@{}} 
\texttt{add}(46 at.\% O) \\
\texttt{settemp}(2200 K) \\ 
\texttt{wait}(60 min)
\end{tabular} 
&
Requires multiple target phases and temperature change. \\
\hline
3
& 
\begin{tabular}{@{}l@{}} 
$T_0 = 300$ K \\
100\% \ch{Nb} 
\end{tabular} 
&
\begin{tabular}{@{}l@{}} 
$T_f = 300$ K \\
50\% \ch{NbO2}, 
50\% \ch{Nb2O5}
\end{tabular} 
&
\begin{tabular}{@{}l@{}} 
\texttt{add}(58 at.\% O) \\
\texttt{settemp}(1300 K) \\ 
\texttt{wait}(60 min) \\
\texttt{settemp}(300 K)
\end{tabular} 
&
Requires multiple target phases that are ``non-adjacent'' to starting phase. \\
\hline
4
& 
\begin{tabular}{@{}l@{}} 
$T_0 = 300$ K \\
100\% \ch{Nb2O5} 
\end{tabular} 
&
\begin{tabular}{@{}l@{}} 
$T_f = 300$ K \\
100\% \ch{NbO} 
\end{tabular} 
&
\begin{tabular}{@{}l@{}} 
\texttt{add}(22 at.\% Nb) \\
\texttt{settemp}(1300 K) \\ 
\texttt{wait}(60 min) \\
\texttt{settemp}(300 K)
\end{tabular} 
&
Single oxide phase that is ``non-adjacent.'' \\
\hline
\end{tabular}
\end{table}

\subsection{Additional results}

\begin{figure}[ht]
\centering
\includegraphics[width=.8\textwidth]{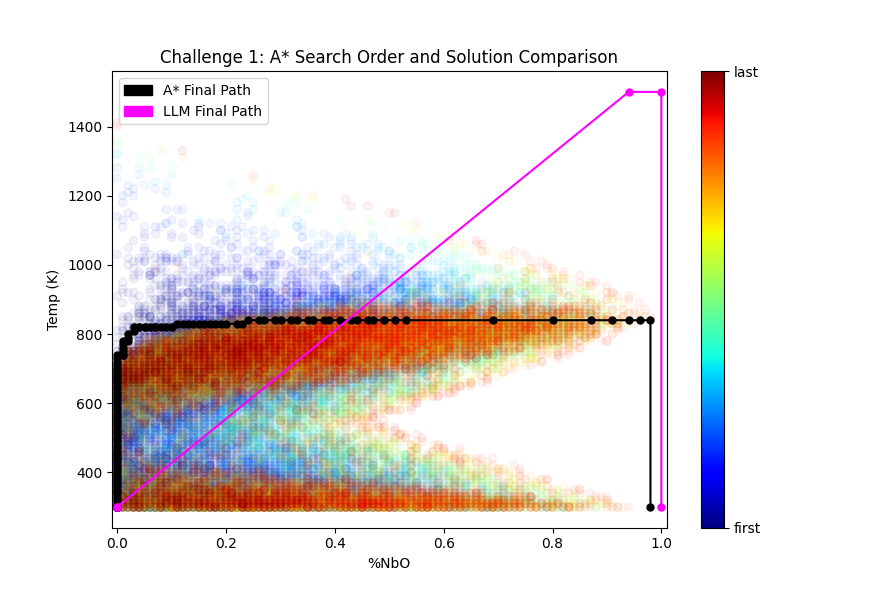}
\caption{Progress of \astar search on the first challenge, projected onto the temperature dimension and the percent NbO dimension (the goal state is 100 \% NbO at 300 K). Marker color denotes order in which states were explored with blue being earliest and red the latest. The final synthesis path of \astar is in black, while the best plan found by the LLM for this challenge is in magenta. 
\label{fig:astarviz}
}
\end{figure}

\Cref{fig:astarviz} provides a visualization of the states explored by one invocation of the \astar algorithm.

\Cref{tbl:llm-solutions-detailed} provides examples of successful synthesis pathways identified by the \gls{LLM} for each challenge.
Through pretraining knowledge, prompted hints, and strategies uncovered by the ReAct technique, the LLM is able to attempt synthesis pathways that are potentially much closer to viable than those prioritized by \astarNS, even though they may appear quite roundabout when plotted in some choice of two dimensions. For example, successful synthesis in the first challenge involves temporarily raising temperature and later lowering it, which is hard to influence via an \astar heuristic but straightforward to convey to an LLM, or even for an LLM with pretraining knowledge to consider without user interventions.

\begin{table}[th]
\centering
\caption{Best synthesis pathways found by the LLM for each challenge.}
\label{tbl:llm-solutions-detailed}
\begin{tabular}{r l p{1.7in}<{\raggedright}}
Id & Best LLM Solution & Comments \\
\hline
1  
&
\begin{tabular}{@{}l@{}} 
\texttt{add}(50 at.\% O), \texttt{settemp}(1500 K), \texttt{wait}(30 min), \\
\texttt{settemp}(300 K), \texttt{wait}(10 min)
\end{tabular} 
& Exactly solves challenge 1. \\
\hline
2
&
\begin{tabular}{@{}l@{}} 
\texttt{add}(47 at.\% O), \texttt{settemp}(1000 K), \texttt{wait}(70 min), \\
\texttt{settemp}(1500 K), \texttt{wait}(70 min), \texttt{settemp}(2200 K),\\ 
\texttt{wait}(10 min)
\end{tabular} 
& Achieves correct temperature but 61\% NbO, 39\% liquid instead of the goal: 49\% NbO, 51\% liquid.\\
\hline
3
&
\begin{tabular}{@{}l@{}} 
\texttt{add}(65 at.\% O), \texttt{settemp}(1700 K), \texttt{wait}(120 min), \\
\texttt{settemp}(1900 K), \texttt{wait}(120 min), \texttt{settemp}(300 K), \\ 
\texttt{wait}(3 min)
\end{tabular} 
& Achieves correct temperature but far from correct material phases.\\
\hline
4
&
\begin{tabular}{@{}l@{}} 
\texttt{add}(50 at.\% Nb), \texttt{settemp}(2000 K), \texttt{wait}(240 min), \\
\texttt{settemp}(300 K), \texttt{wait}(10 min)
\end{tabular} 
& Exactly solves challenge 4. \\
\hline
\end{tabular}
\end{table}

\clearpage
\subsection{Additional LLM details}
The \gls{LLM}'s prompt consists of a context string, which provides the bulk of the detail, and the challenge problem itself, which defines the start and goal states. The context string is shown in~\cref{fig:llmcontext}.
For all of our runs we used a \gls{LLM} temperature setting of 0.1.

\begin{figure}[h]
\begin{tcolorbox}[colback=gray!10,colframe=black!50,sharp corners,label=lst:llmprompt,title=LLM Prompt,fontupper=\small]
  You should consider raising temperatures or holding them over long periods (hours), as well as fast temperature changes (seconds).
  Even if you get close, try a few more options to see if you can get closer. \\

  Action sequences are specified as a list of dictionaries, with one dictionary for each action.
  There are three types of actions: temperature change actions, composition change actions, and wait actions. \\
 
  * Temperature change actions increase or decrease temperature over the provided amount of time. \\
  * Composition change actions introduce raw precursors (Nb or O) in order to realize the specific percentages. Note the arguments to this action are the final percentages to realize, NOT the amount to be added. \\
  * Wait actions specify a duration of time where there is no change to composition or temperature. This may be needed (with a large duration) if you want to approach thermal equilibrium. \\

  Valid keys for action dictionary are "Nb", "O", "Temperature", and "Time", representing the fraction of niobium in the overall composition,
  the fraction of oxygen in the overall composition, the temperature, and the time taken from the last step.
  \vspace{1em}

  * Composition entries may be omitted if they can be inferred from the implied sum-to-1 constraint.  \\
  * Temperature actions should not specify any raw material changes; these must be changed separately. \\
  * Use your knowledge of materials science and the underlying phase diagram when devising the plan. A successful plan may not necessarily be a straightforward trajectory. \\
  * All times are in units of seconds and all temperatures are in units of Kelvin. \\
  * Feel free to try as many examples as you like in the process of searching for a solution. Don't give up until you can't think of any other ways to try to solve the problem. \\

  Example: (1) raise temperature to 1500K over 10 seconds, (2) wait for 60 minutes, then (3) add precursors so that the fraction of oxygen is 80\% of the mixture: \\

  [\{"Temperature":1500, "Time":10.0\}, \{"Wait": 3600\}, \{"O":0.8, "Nb":0.2\}] \\

  Example: (1) add raw precursors achieve 70\% Nb then (2) heat to 2100K over 5 seconds, and finally (3) cool back down to 300K over one minute: \\
  
  [\{"Nb":0.7\}, \{"Temperature":2100, "Time":5.0\}, \{"Temperature":300, "Time":60.0\}] \\

  Your final answer must be a synthesis plan in the above form. You must provide a synthesis plan, even if you don't think it will solve the problem correctly.
\end{tcolorbox}
\caption{LLM context string.} 
\label{fig:llmcontext}
\end{figure}


\begin{figure}[h]
\begin{tcolorbox}[colback=gray!10,colframe=black!50,sharp corners,label=lst:llmprompt2,title=Example LLM Output on Challenge 4 (Successful),fontupper=\small]

\textbf{Generate a synthesis plan for the following problem:}

\textbf{Start state:} 
TEMPERATURE: 300, COMPOSITION: [NB: 0.29, O: 0.71], PHASE FRACTIONS: [NB2O5: 1.0].

\textbf{Goal state:}
TEMPERATURE: 300, PHASE FRACTIONS: [NBO: 1.0].
\\

\textbf{\textit{... LLM Context inserted here ...}}
\\

\textbf{Thought 1:} To transform Nb2O5 into NBO, we need to consider the phase transformations that occur with temperature changes. Nb2O5 can be reduced to NBO by controlling the temperature and possibly the composition. Initially, I will try a synthesis plan that involves heating the material to a high temperature to facilitate the reduction and then cooling it back to the desired temperature. I will start by heating to 1500K over 10 seconds, wait for 3600 seconds to allow the reaction to occur, and then cool back to 300K over 60 seconds.

\textbf{Tool 1:} \textit{simulate([Temperature: 1500, Time: 10.0], [Wait: 3600], [Temperature: 300, Time: 60.0])}

\textbf{Observation 1:} \textit{final\_temp=300; final\_phases=[NB2O5: 0.9099999999999995, BCC\_A2: 0.0, NBO: 0.0, HCP\_A3: 0.0, GAS: 0.0, ION: 0.0, LIQUID: 0.0, NBO2: 0.09000000000000052, FCC\_A1: 0.0]}

\textbf{Thought 2:} The initial attempt did not produce the desired NBO phase... \textit{(truncated)}. \\

\textbf{\textit{... Truncated for Readability ...}}
\\

\textbf{Thought 4:} The third attempt also failed to produce the desired NBO phase, with the majority of the material remaining as Nb2O5 and a small amount converting to NBO2. This suggests that the conditions are still not optimal for the conversion to NBO. I will try a different strategy by adjusting the composition to increase the niobium content, which might facilitate the formation of NBO. I will add raw precursors to achieve 50\% Nb, then heat to 1700K over 10 seconds, wait for 7200 seconds, and finally cool back to 300K over 60 seconds.

\textbf{Tool 4:} \textit{simulate([Nb: 0.5, O: 0.5], [Temperature: 1700, Time: 10.0], [Wait: 7200], [Temperature: 300, Time: 60.0])}

\textbf{Observation 4:} \textit{final\_temp=300; final\_phases={NB2O5: 0.0, BCC\_A2: 2.2024444095574645e-06, NBO: 0.9999917981555904, HCP\_A3: 0.0, GAS: 0.0, ION: 0.0, LIQUID: 0.0, NBO2: 5.9994000001725e-06, FCC\_A1: 0.0}}

\textbf{Thought 5:} The final attempt successfully produced the desired NBO phase with a phase fraction of approximately 0.99999. This indicates that the strategy of adjusting the composition to increase the niobium content, followed by heating to 1700K and holding for 7200 seconds, was effective in achieving the goal state. The synthesis plan is now complete.\\

\textbf{Response:}
The synthesis plan to convert Nb2O5 to NBO involved several attempts to find the optimal conditions for the phase transformation. Initially, attempts to simply heat the material to high temperatures and wait did not yield the desired NBO phase. This indicated that the temperature and duration were not sufficient or appropriate for the conversion. Adjusting the composition to increase the niobium content proved to be a successful strategy. By achieving a 50\% Nb composition, heating to 1700K, and holding for 7200 seconds, the desired NBO phase was successfully formed with a phase fraction of approximately 0.99999. This approach balanced the composition and thermal conditions to facilitate the desired phase transformation.

\textbf{Answer:}
[Nb: 0.5, O: 0.5], [Temperature: 1700, Time: 10.0], [Wait: 7200], [Temperature: 300, Time: 60.0]

\end{tcolorbox}
\caption{Successful attempt on challenge 4 by the LLM using the ReAct framework. The output has been manually formatted for readability and to fit the page. The plan produced by the LLM achieves the correct phase and the correct final temperature.} 
\label{fig:llmoutputsuccess}
\end{figure}


\begin{figure}[h]
\begin{tcolorbox}[colback=gray!10,colframe=black!50,sharp corners,label=lst:llmprompt3,title=Example LLM Output on Challenge 4 (Unsuccessful),fontupper=\small]

\textbf{Generate a synthesis plan for the following problem:}

\textbf{Start state:} 
TEMPERATURE: 300, COMPOSITION: [NB: 0.29, O: 0.71], PHASE FRACTIONS: [NB2O5: 1.0].

\textbf{Goal state:}
TEMPERATURE: 300, PHASE FRACTIONS: [NBO: 1.0].
\\

\textbf{\textit{... LLM Context inserted here ...}}
\\

\textbf{Thought 1:} To transform Nb2O5 into NBO, we need to consider the phase transformations and the necessary conditions for these transformations. Typically, reducing Nb2O5 to NBO involves a reduction process, which might require a high temperature to facilitate the reaction. We should start by increasing the temperature significantly to see if this promotes the desired phase change. After heating, we can hold the temperature to allow the system to reach equilibrium. Let's start by heating the system to a high temperature and then holding it for a while.

\textbf{Tool 1:} \textit{simulate([Temperature: 1800, Time: 60], [Wait: 7200])}

\textbf{Observation 1:} \textit{final\_temp=1800; final\_phases=[NB2O5: 0.07840216105964366, GAS: 0.0, NBO: 0.0, LIQUID: 0.0, NBO2: 2.0829399531378196e-05, BCC\_A2: 0.0, FCC\_A1: 0.0, ION: 0.9215770095408249, HCP\_A3: 0.0]}

\textbf{Thought 2:} The initial attempt to heat the system to 1800K and hold it for 2 hours did not result in the formation of NBO.... \textit{(truncated)}. \\

\textbf{\textit{... Truncated for Readability ...}}
\\

\textbf{Thought 4:} The stepwise temperature increase with intermediate holds still resulted in the complete conversion to the ION phase, with no formation of NBO. This suggests that the current approach of only adjusting temperature is not effective in achieving the desired phase transformation. We may need to consider altering the composition to favor the formation of NBO. Let's try adding Nb to increase its proportion in the composition and see if this helps in forming NBO.

\textbf{Tool 4:} \textit{simulate([Nb: 0.5], [Temperature: 1500, Time: 60], [Wait: 3600], [Temperature: 1800, Time: 60], [Wait: 7200], [Temperature: 2100, Time: 60], [Wait: 14400])}

\textbf{Observation 4:} \textit{final\_temp=2100; final\_phases=[NB2O5: 0.0, GAS: 0.0, NBO: 0.9999889654813701, LIQUID: 0.0, NBO2: 0.0, BCC\_A2: 2.3775424105629224e-06, FCC\_A1: 0.0, ION: 8.656976219446085e-06, HCP\_A3: 0.0]}

\textbf{Thought 5:} The addition of Nb to increase its proportion in the composition, followed by a stepwise temperature increase, successfully resulted in the formation of NBO with a phase fraction of approximately 0.99999. This indicates that the synthesis plan is effective in achieving the desired phase transformation from Nb2O5 to NBO. The final synthesis plan involves adjusting the composition and carefully controlling the temperature profile.\\

\textbf{Response:}
The initial attempts to transform Nb2O5 into NBO by only adjusting the temperature were unsuccessful, as they resulted in the formation of the ION phase instead. This indicated that the conditions were not suitable for the desired phase transformation. By increasing the proportion of Nb in the composition, the conditions became favorable for the formation of NBO. The successful synthesis plan involved adding Nb to adjust the composition, followed by a stepwise temperature increase with intermediate holds. This approach allowed the system to reach the desired phase transformation, achieving a phase fraction of NBO close to 1.0.

\textbf{Answer:}
[Nb: 0.5], [Temperature: 1500, Time: 60], [Wait: 3600], [Temperature: 1800, Time: 60], [Wait: 7200], [Temperature: 2100, Time: 60], [Wait: 14400]

\end{tcolorbox}
\caption{Unsuccessful attempt on challenge 4 by the LLM using the ReAct framework. The output has been manually formatted for readability and to fit the page. While this plan achieves the desired phase, it forgets to account for the goal temperature and ends with the material heated to 2100K.} 
\label{fig:llmoutputfail}
\end{figure}

\end{document}

%% file: math_commands.tex
\usepackage{amsmath,amsfonts,bm}

\def\eqref#1{equation~\ref{#1}}

\def\1{\bm{1}}

\DeclareMathAlphabet{\mathsfit}{\encodingdefault}{\sfdefault}{m}{sl}
\SetMathAlphabet{\mathsfit}{bold}{\encodingdefault}{\sfdefault}{bx}{n}

